\documentclass[review]{elsarticle}

\usepackage{times}
\usepackage{soul}
\usepackage{url}
\usepackage[hidelinks]{hyperref}
\usepackage{graphicx}
\usepackage{amsmath}
\usepackage{amsthm}
\usepackage{booktabs}
\usepackage{algorithm}
\usepackage{algorithmic}
\usepackage{amsfonts,amssymb}
\usepackage{color}
\usepackage{parskip}
\usepackage{caption}
\usepackage{subcaption}
\usepackage[section]{placeins}
\usepackage{multirow}
\usepackage{todonotes}

\journal{Journal of \LaTeX\ Templates}

\begin{document}

\begin{frontmatter}

\title{Unveiling the Potential of Spiking Dynamics in Graph Representation Learning through Spatial-Temporal Normalization and Coding Strategies}
\tnotetext[mytitlenote]{Corresponding author,$^\dagger$Contribute equally}


\author[mymainaddress,mysecondaryaddress]{Mingkun Xu $^\dagger$}
\author[mysecondaryaddress]{Huifeng Yin $^\dagger$}
\author[address3]{Yujie Wu}
\author[address4]{Guoqi Li}
\author[mysecondaryaddress]{Faqiang Liu}
\author[mysecondaryaddress]{Jing Pei}
\author[mymainaddress]{Shuai Zhong $^{\ast}$}
\author[mysecondaryaddress]{Lei Deng $^{\ast}$}

\address[mymainaddress]{Guangdong Institute of Intelligence Science and Technology, Hengqin, Zhuhai, China}
\address[mysecondaryaddress]{Center for Brain-Inspired Computing Research (CBICR),Tsinghua University, Beijing, China}
\address[address3]{Hong Kong Polytechnic University, Hong Kong, China}
\address[address4]{Institute of Automation, Chinese Academy of Sciences, Beijing, China}

\begin{abstract}
In recent years, spiking neural networks (SNNs) have attracted substantial interest due to their potential to replicate the energy-efficient and event-driven processing of biological neurons. Despite this, the application of SNNs in graph representation learning, particularly for non-Euclidean data, remains underexplored, and the influence of spiking dynamics on graph learning is not yet fully understood. This work seeks to address these gaps by examining the unique properties and benefits of spiking dynamics in enhancing graph representation learning. We propose a spike-based graph neural network model that incorporates spiking dynamics, enhanced by a novel spatial-temporal feature normalization (STFN) technique, to improve training efficiency and model stability. Our detailed analysis explores the impact of rate coding and temporal coding on SNN performance, offering new insights into their advantages for deep graph networks and addressing challenges such as the oversmoothing problem. Experimental results demonstrate that our SNN models can achieve competitive performance with state-of-the-art graph neural networks (GNNs) while considerably reducing computational costs, highlighting the potential of SNNs for efficient neuromorphic computing applications in complex graph-based scenarios.

\end{abstract}


\end{frontmatter}


\section{Introduction}
Spiking neural networks (SNNs) are brain-inspired computational models that replicate the spatio-temporal dynamics and rich coding schemes inherent in biological neural systems. These features make SNNs particularly adept at mimicking neuroscience-inspired models and performing efficient computations on neuromorphic hardware. SNNs have been successfully applied to a variety of tasks, including image classification, voice recognition, object tracking, and neuromorphic perception \cite{wu2018spatio, silva2017evolving, cao2015spiking, zhu2020retina, haessig2019spiking,li2023brain}. However, most existing studies have focused on processing unstructured data in Euclidean space, such as images and language \cite{zhou2018graph}. Meanwhile, biological spiking neurons, characterized by their complex coding schemes and dynamic properties, are foundational in creating cognitive maps in the hippocampal-entorhinal system. These maps are crucial for organizing and integrating spatial relationships and relational memory, facilitating advanced cognitive functions \cite{whittington2020tolman}. Thus, there is a strong motivation to explore effective algorithms for training SNNs that can integrate topological relationships and structural knowledge from graph-based data.

In recent years, there have been attempts to combine SNNs with graph-based scenarios, but these have primarily focused on applying graph theory to analyze spiking neuron characteristics and network topologies \cite{piekniewski2007emergence, cancan2019ev, jovanovic2016interplay}, or utilizing spiking neuron features to tackle simple graph-related problems such as shortest path, clustering, and minimal spanning tree problems \cite{sala1999solving, hamilton2020spike}. A recent study \cite{gu2020tactilesgnet} introduced graph convolution to preprocess tactile data for SNN classification, achieving high performance in sensor data classification. However, this approach faces challenges in accommodating general graph operations and adapting to other scenarios. In addition, although some works have investigated the modelling of SNN and graph learning \cite{zhu2022spiking,yin2024dynamic,li2023scaling}, they overlooked a more in-depth and systematic exploration of how spiking dynamics could impact and enhance graph representation learning.
On the other hand, numerous graph neural network (GNN) models, such as Graph Convolution Network (GCN) and Graph Attention Network (GAT), have made significant advances in solving graph-related tasks \cite{kipf2019semi, velivckovic2017graph}. Despite these advancements, few works have explored the interplay between graph theory and neural dynamics, and many suffer from substantial computational overhead when applied to large-scale data, potentially limiting their practical applications.

This study aims to address these challenges by investigating the potential of spiking dynamics in graph representation learning, focusing on the development of a comprehensive spike-based modeling framework. We also introduce spatial-temporal feature normalization techniques to enhance training efficiency and model stability. Our approach leverages the unique benefits of SNNs in processing structured data, providing insights into reducing computational costs and improving stability in graph-based learning scenarios.

To investigate the potential and benefits of spiking dynamics in the context of graph-structured data, we introduce an innovative graph SNN framework designed to handle non-Euclidean data using flexible aggregation methods. The development of such SNN models presents two major challenges. The first is integrating multiple graph convolution operations with spiking dynamics. To address this challenge, we unfold binary node features across both temporal and spatial dimensions, proposing a comprehensive spiking message-passing technique that integrates graph filters with spiking dynamics iteratively.
This approach allows for the effective merging of graph convolution operations and spiking dynamics, leveraging the unique properties of SNNs for processing complex, structured information. By iteratively incorporating graph filters into the spiking message-passing process, our framework can dynamically adapt to the intricate data structures typical of non-Euclidean spaces, enhancing the model's ability to abstract and generalize features from graph-based input
The second major challenge is ensuring the convergence and performance of training in graph SNNs. Given the intricate spiking dynamics and the diverse nature of graph operations, directly training SNNs on graph-structured data remains an underexplored and challenging area. While various normalization techniques have proven effective for enhancing network convergence \cite{ioffe2015batch,ba2016layer}, their direct application to graph SNNs is hindered by the unique challenges posed by temporal neuronal dynamics and the binary communication mechanisms inherent in SNNs \cite{zheng2020going}. Thus, there is a pressing need for a novel normalization method tailored to the specific requirements of spiking dynamics in graph scenarios.
To address this issue, we propose a spatial-temporal feature normalization (STFN) algorithm. This method normalizes the instantaneous membrane potentials across both feature and temporal dimensions for each node, thereby enhancing the SNN's ability to extract latent features from aggregated signals within a graph.
Our contributions can be summarized as follows:
\begin{itemize}
\item We have developed a comprehensive spike-based modeling framework, termed Graph SNNs, which integrates spike propagation with feature affine transformations, effectively reconciling graph convolution operations with spiking communication mechanisms. This framework is highly adaptable and supports a wide range of graph propagation operations. To our knowledge, this is the first framework to establish a general spike-based approach for regular graph tasks within a gradient-descent paradigm.
\item We introduce the spatial-temporal feature normalization (STFN) algorithm, which accounts for temporal neuronal dynamics and aligns membrane potential representations with threshold levels, significantly improving convergence and performance.
\item We have instantiated our proposed framework into multiple models, including Graph Convolutional SNNs (GC-SNN) and Graph Attention SNNs (GA-SNN), and validated their performance through extensive experiments on semi-supervised node classification tasks. Additionally, we evaluate the computational costs, demonstrating the high efficiency and potential of SNNs for enabling graph-structured applications on neuromorphic hardware.
\end{itemize}

\section{Related Work}
\subsection{Spiking Neural Networks}
Over the past few years, Spiking Neural Networks (SNNs) have gained significant attention in the field of neuromorphic computing, which seeks to emulate the biological plausibility of the human brain. Unlike traditional Artificial Neural Networks (ANNs), SNNs are adept at processing sequential data. These networks mimic the behavior of biological neurons by incorporating synaptic and neuronal dynamics, utilizing binary spikes that are either 0 or 1. Among the various SNN models, the Leaky Integrate-and-Fire (LIF)\cite{gerstner2014neuronal} and Hodgkin-Huxley (HH)\cite{hodgkin1952quantitative} models are notably prominent. Due to the high computational demand of the HH model, the LIF model is more commonly adopted, as it offers a practical compromise between accuracy and computational efficiency. The LIF model can be mathematically described as follows:

\begin{equation}
\tau \frac{d u(t)}{d t}=-u(t)+I(t)
\end{equation}
In this context, \(u(t)\) denotes the membrane potential of the neuron at a given time \(t\), \(I(t)\) signifies the external input current, and \(\tau\) represents the membrane time constant, which influences the rate of potential decay.

Inspired by the complex and varied connectivity observed in the brain, spiking neural networks (SNNs) can be organized into various architectures. These include fully-connected networks\cite{gutig2016spiking}, liquid state machines\cite{maass2011liquid}, spiking convolutional neural networks\cite{cao2015convolutional}, and spiking Boltzmann machines\cite{neftci2014event}, among others. Such diverse structures enable SNNs to efficiently and effectively process intricate, sparse, and noisy spatio-temporal information. Despite their potential, training SNNs poses significant challenges due to the non-differentiable nature of spike-based communication. Currently, there are three primary methods for training SNNs\cite{wu2018spatio}: unsupervised learning (such as spike-timing-dependent plasticity (STDP)\cite{wu2018spatio}), conversion learning (also known as the ANN-to-SNN conversion method), and supervised learning (which employs gradient descent to train multi-layer SNNs, achieving accuracy comparable to ANNs on complex datasets\cite{wu2018spatio}).


In general, Spiking Neural Networks (SNNs) have not been widely adopted for graph learning. However, recent research is increasingly focusing on this approach. For instance, Zhu et al.\cite{zhu2022spiking} integrated Graph Convolutional Networks (GCNs) with the biologically inspired properties of SNNs, significantly enhancing efficiency while achieving strong performance on graph datasets. Similarly, Gu et al.\cite{gu2020tactilesgnet} structured tactile data into a graph format and leveraged the event-driven characteristics of SNNs, attaining a high classification accuracy of around 90

\subsection{Graph Neural Networks}
Graphs are common data structures used to represent complex systems, consisting of vertices (nodes) that symbolize entities and edges that indicate relationships between these nodes. Depending on the nature of these relationships, graphs can be classified into several categories, such as directed or undirected, weighted or unweighted, and cyclic or acyclic. Graph Neural Networks (GNNs) are specialized models designed to work with graph data. They leverage both the features of the nodes and the graph's structural information to learn from complex graph datasets. GNNs can be applied to various tasks, including node-level tasks (predicting properties for individual nodes) and graph-level tasks (predicting a single property for the entire graph).

The architecture of Graph Neural Networks (GNNs) can be broadly classified into four categories: Convolutional Graph Neural Networks (GCNs)\cite{niepert2016learning, gilmer2017neural}, Graph Attention Networks (GATs)\cite{velickovic2017graph}, Graph Autoencoders (GAEs)\cite{kipf2016variational}, and Spatial-Temporal Graph Neural Networks (STGNNs)\cite{yu2017spatio}. GCNs are among the most widely used GNN models, employing convolution operations on graphs to aggregate information from neighboring nodes. These can be further divided into spectral versions, which operate in the spectral domain, and spatial versions, which work directly within the graph's node space. GATs utilize attention mechanisms to prioritize the influence of neighboring nodes, allowing the model to focus on the most relevant information during the learning process. GAEs are unsupervised learning models that use an encoder to project input graph data into a low-dimensional vector space and a decoder to reconstruct the graph or specific properties from this space. Finally, STGNNs are designed to manage graph data with both spatial and temporal aspects, making them particularly useful for tasks such as traffic forecasting, human motion prediction, and other analyses involving time-series data on graphs.

However, all these models encounter the over-smoothing problem as the depth of a GNN increases\cite{zhou2021understanding}. During the training phase, with multiple layers of message passing, the representations of nodes tend to become more similar to one another. This convergence can result in the loss of important information and reduce the model's ability to distinguish between different nodes. This challenge is a major obstacle in designing deep GNNs.

\subsection{Normalization}
Normalization has been crucial in the training of deep learning models, enhancing training efficiency and mitigating issues like vanishing or exploding gradients and internal covariate shift. By scaling data with varying distributions to a consistent range, normalization also speeds up the model's convergence. A pivotal advancement in this area was Batch Normalization (BN)\cite{ioffe2015batch}, introduced by Ioffe and Szegedy, which normalizes activations within a mini-batch during training, thus accelerating convergence. BN works by adjusting the output of each neuron by subtracting the batch mean and dividing by the batch standard deviation. Subsequent techniques like Layer Normalization (LN)\cite{ba2016layer} and Group Normalization (GN)\cite{wu2018group} have emerged as effective regularizers, helping to reduce overfitting and enhance the generalization capabilities of models.

In graph neural networks (GNNs), adapting normalization techniques poses unique challenges due to the non-Euclidean nature of the data and the complex relationships between entities. Layer normalization, tailored for GNNs, helps stabilize training and enhance convergence\cite{velickovic2017graph}. Normalizing the adjacency matrix using the degree matrix, known as adjacency matrix normalization, is another crucial step that ensures scale-invariant information propagation\cite{kipf2019semi}. Additionally, spectral normalization, which controls the Lipschitz constant of the network, has been applied to GNNs to prevent overfitting to specific graph structures or nodes\cite{miyato2018spectral}. To address the over-smoothing issue and simultaneously increase GNN depth, Zhou et al. introduced the NodeNorm method\cite{zhou2021understanding}. This method scales the hidden features of GNNs based on the standard deviation of each individual node, making the normalization effect controllable and compatible with GCNs in various scenarios.

Despite the development of these normalization techniques for GNNs, they have not been effectively applied to Graph Spiking Neural Networks. This gap represents a significant challenge and offers a promising research opportunity.


\begin{figure*}[h]
\begin{center}
\includegraphics[width=12.5cm]{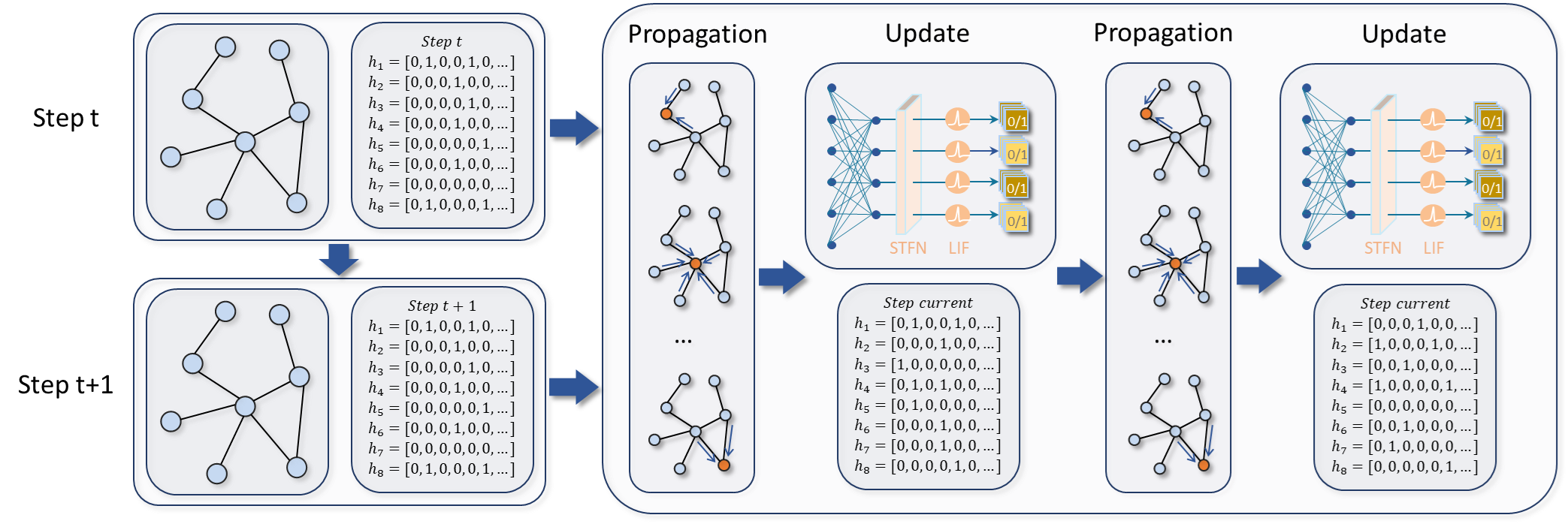} 
\end{center}
\caption{The Graph SNN framework can support spiking message propagation and feature affine transformation, reconciling the graph convolution operation and spiking communication mechanism in a unified paradigm. The proposed STFN normalizes the membrane potentials along both spatial and temporal dimension, which coordinates the data distribution with threshold but also facilitates the network convergence.
}
\label{architecture}
\end{figure*}

\section{Graph Spiking Neural Networks}
We present our Graph SNN framework, focusing on the following four key aspects: (1) a method for integrating graph convolution operations into the spiking computing paradigm; (2) an iterative spiking message passing model, along with an overall training framework that consists of two phases; (3) a spatial-temporal feature normalization technique to facilitate convergence; and (4) implementations of specific spiking graph models.

\subsection{Spiking Graph Convolution}
Given an attributed graph \( \mathcal{G} = (\mathcal{V}, \mathcal{E}) \), where \( \mathcal{V} \) represents the set of nodes and \( \mathcal{E} \) represents the set of edges, the graph attributes are typically described by an adjacency matrix \( A \in \mathbb{R}^{N \times N} \) and a node feature matrix \( X \in \mathbb{R}^{N \times C} = [x_1; x_2; \ldots; x_N] \). Here, each row \( x_i \) of \( X \) represents the feature vector of a node, and the adjacency matrix \( A \) satisfies \( A_{ii} = 0 \), meaning there are no self-loops. If the signal of a single node is represented by a feature vector \( x \in \mathbb{R}^C \), the spectral convolution is defined as the multiplication of a filter \( \mathcal{F}_\theta = \text{diag}(\theta) \) with the node signal \( x \), expressed as:
\begin{equation}
 \mathcal{F}_\theta \star x = U \mathcal{F}_\theta U^{\top} x,
\end{equation}
where \(\mathcal{F}_\theta\) is parameterized by \(\theta \in \mathbb{R}^C\), and \(U\) is the eigenvector matrix of the Laplacian matrix \(L = I - D^{-\frac{1}{2}}AD^{-\frac{1}{2}} = U\Lambda U^{\top}\). Here, \(U^{\top} x\) can be interpreted as the graph Fourier transform of the signal \(x\), and the diagonal matrix \(\Lambda\), containing the eigenvalues, can be filtered using the function \(\mathcal{F}_\theta(\Lambda)\). However, due to the high computational cost associated with this approach, some methods have proposed using approximations and stacking multiple nonlinear layers to reduce the overhead, which has been successfully implemented in recent work\cite{kipf2019semi}.

To unify the flow of information within spiking dynamics, we convert the initial node signals \( X \) into binary components \( \{\widetilde{X}_0, \widetilde{X}_1, \ldots, \widetilde{X}_t, \ldots, \widetilde{X}_{T-1}\} \), where \( T \) represents the length of the time window. The encoding process can be either probabilistic, following distributions such as Bernoulli or Poisson, or deterministic, involving methods like quantitative approximation or the use of an encoding layer to produce spikes\cite{wu2019direct}. This approach transforms the attributed graph from its original state into a spiking representation. We denote the spiking node embedding at time \( t \) in the \( n \)-th layer as \(\widetilde{H}_t^n\), with \(\widetilde{H}_t^0 = \widetilde{X}_t\) (where the tilde indicates binary variables in spike form). The layer-wise spiking graph convolution in the spatial-temporal domain is then defined as:
\begin{equation}
\widetilde{H}_t^{n} =  \Phi (\mathcal{G}_c(A, \widetilde{H}_t^{n-1}) W^n, \widetilde{H}_{t-1}^{n}),
\label{spi-conv}
\end{equation}
In this context, \(\mathcal{G}_c(A, \widetilde{H})\) represents the spiking feature propagation along the graph's topological structure, which can be implemented using various propagation methods, such as the Chebyshev filter or the first-order model. \(\Phi(\cdot)\) denotes the non-linear dynamic process that depends on historical states \(\widetilde{H}_{t-1}^{n}\). The matrix \(W^n \in \mathbb{R}^{C^{n-1} \times C^n}\) is a layer-specific trainable weight parameter, where \(C^n\) indicates the output dimension of the spiking features in the \(n\)-th layer, with \(C^0 = C\) being the input feature dimension. This formula describes a single spiking graph convolution layer, and a stacked multi-layer structure can create a comprehensive system for modeling spiking dynamics.

\subsection{Iterative Spiking Message Passing}
We use the leaky integrate-and-fire (LIF) model as our fundamental neuron unit. This model is computationally efficient and widely used, while also preserving a certain degree of biological realism. The dynamics of the LIF neuron can be described by:
\begin{equation}
\label{lif}
\tau \frac{dV(t)}{dt} = -(V(t)-V_{reset}) + I(t),
\end{equation}
In this model, \(V(t)\) represents the membrane potential of the neuron at time \(t\). The parameter \(\tau\) is the time constant, and \(I(t)\) denotes the pre-synaptic input, which is the result of synaptic weights combined with the activities of pre-synaptic neurons or external stimuli. When \(V(t)\) exceeds a certain threshold \(V_{th}\), the neuron generates a spike and the membrane potential is reset to \(V_{reset}\). After emitting a spike, the neuron begins to accumulate membrane potential \(V(t)\) again in the following time steps.

The spiking message passing process comprises an information propagation step and an update step, both of which occur over \(T\) time steps. Let \(\widetilde{H}_t = [\widetilde{h}_t^0; \widetilde{h}_t^1; \ldots; \widetilde{h}_t^{N-1}] \in \mathbb{R}^{N \times C}\) represent the node embeddings at time \(t\), where each \(\widetilde{h}_t^i\) corresponds to the feature vector of node \(i\). For a node \(v\), we provide a general formulation for its spiking message passing as:
\begin{equation}
\label{mpss}
\widetilde{h}^v_t = U(\sum_{u \in N(v)}P(\widetilde{h}^v_t, \widetilde{h}^u_t, e_{vu}),\widetilde{h}^v_{t-1})),
\end{equation}
In this context, \(P(\cdot)\) denotes the spiking message aggregation from neighboring nodes, which can be implemented using various graph convolution operations, represented as \(\mathcal{G}_c(\cdot)\). The function \(U(\cdot)\) signifies the state update, governed by a non-linear dynamic system. \(N(v)\) represents the set of all neighbors of node \(v\) in the graph \(\mathcal{G}\), and \(e_{vu}\) indicates the static edge connection between nodes \(v\) and \(u\), which can be naturally extended to the formalism of directed multigraphs. Equation \((\ref{spi-conv})\) provides a specific implementation of this general formulation.

To integrate the LIF model into the above framework, we employ the Euler method to convert the first-order differential equation in Eq. \((\ref{lif})\) into an iterative form. We introduce a decay factor \(\kappa\) to represent the term \((1 - \frac{dt}{\tau})\) and express the pre-synaptic input \(I\) as \(\sum_j W^j \mathcal{G}_c(A, \widetilde{H}^j_{t+1})\). Here, graph convolution is used to implement the propagation step \(P(\cdot)\). Incorporating the scaling effect of \(\frac{dt}{\tau}\) into the weight term, we derive the following formulation:
\begin{equation}
V_{t+1} = \kappa V_t + \sum_j W^j\mathcal{G}_c(A, \widetilde{H}^j_{t+1}).
\end{equation}
\(\mathcal{G}_c(A, \widetilde{H}^j_{t+1})\) represents the aggregated feature from the pre-synaptic neurons, with the superscript \(j\) indicating the index of the pre-synapse. By incorporating the firing-and-resetting mechanism and assuming \(V_{reset} = 0\), the update equation can be expressed as follows:
\begin{align} 
&  V^{n,i}_{t+1} = \kappa V^{n,i}_t(1-\widetilde{H}^{n,i}_t) + \sum^{l(n)}_j W^{n,ij} \mathcal{G}_c(A, \widetilde{H}^{n-1,j}_{t+1}), \label{update1} \\
& \widetilde{H}^{n,i}_{t+1} = g(V^{n,i}_{t+1} - V_{th}), \label{firing} 
\end{align} 
Here, \(n\) denotes the \(n_{th}\) layer, and \(l(n)\) indicates the number of neurons in that layer. \(W^{ij}\) represents the synaptic weight from the \(j_{th}\) neuron in the pre-layer to the \(i_{th}\) neuron in the post-layer. The function \(g(\cdot)\) is the Heaviside step function, which is used to model the neuron's spiking behavior. By this approach, we transform the implicit differential equation into an explicit iterative form, which can effectively describe the message propagation and update process outlined in equation \((\ref{mpss})\) and illustrated in Figure 1.

Moreover, we observe that the computation order of the affine transformation and graph convolution can be interchanged when the graph convolution operation is linear (e.g., \(\mathcal{G}_c(A, H) = D^{-\frac{1}{2}} AD^{-\frac{1}{2}}H\)). Given that the matrix \(W\) is typically dense, while \(\widetilde{H}\) is sparse and binary, prioritizing the calculation of \(\widetilde{H}\) and \(W\) can reduce computational overhead by converting multiplications into additions. Under these conditions, the process in equation \((\ref{update1})\) can be reformulated as follows:
\begin{equation}
\label{update2}
V^{n,i}_{t+1} = \kappa V^{n,i}_t(1-\widetilde{H}^{n,i}_t) + \mathcal{G}_c(A, \sum^{l(n)}_j W^{n,ij} \widetilde{H}^{n-1,j}_{t+1}).
\end{equation}
In this way, we present a universal spiking message passing framework in an iterative manner. By specifying \(\mathcal{G}_c(\cdot)\), most proposed graph convolution operations\cite{kipf2019semi, gilmer2017neural, hamilton2017inductive} can be incorporated into this model, making it versatile and adaptable to various graph scenarios. This framework provides a cohesive approach for integrating different types of graph convolution operations, facilitating its application across diverse graph-based tasks.

\subsection{Spatial-temporal Feature Normalization}
Due to the inclusion of temporal dynamics and the event-driven nature of spiking binary representation, traditional normalization techniques cannot be directly applied to Spiking Neural Networks (SNNs). Additionally, the direct training of SNNs on graph tasks does not ensure convergence or optimal performance. This challenge has led us to propose a spatial-temporal feature normalization (STFN) algorithm specifically designed to address the spiking dynamics in graph-based scenarios.

Considering the feature map calculation step, let \(S_t \in \mathbb{R}^{N \times C}\) represent the instantaneous membrane potential output of all neurons in a layer at time step \(t\). This can be expressed as \(\sum^{l(n)}_j W^{n,ij} \mathcal{G}_c(A, \widetilde{H}^{n-1,j}_{t})\) as shown in equation \((\ref{update1})\), or as \(\sum^{l(n)}_j W^{n,ij} \widetilde{H}^{n-1,j}_{t}\) as shown in equation \((\ref{update2})\).
In the STFN process, pre-synaptic inputs are normalized along the feature dimension \(C\) independently for each node. Given the importance of temporal effects in transforming node features within the topology space, normalization is also applied along the temporal dimension across consecutive time steps. Let \(S^{k,v}_t\) denote the \(k_{th}\) element in the feature vector of node \(v\) at time \(t\). The normalization of \(S^{k,v}_t\) is performed as follows:
\begin{equation}
\label{stfn}
\begin{aligned} 
& \hat{S}_t^{k,v} = \frac{\rho V_{th}(S_t^{k,v}-E[S^v])}{\sqrt{Var[S^v] + \epsilon}},
\\
& Y_t^{k,v} = \lambda^{k,v} \hat{S}_t^{k,v} + \gamma^{k,v},
\end{aligned} 
\end{equation}
where \(\rho\) is a hyperparameter optimized during the training process, and \(\epsilon\) is a small constant to prevent division by zero. \(\lambda^{k,v}\) and \(\gamma^{k,v}\) are two trainable parameter vectors specific to node \(v\). \(E[S^v]\) and \(Var[S^v]\) represent the mean and variance of the feature values of node \(v\), respectively, computed across the feature and temporal dimensions. Figure 2 illustrates the calculation process of \(E[S^v]\) and \(Var[S^v]\), which are defined as follows:
\begin{equation}
\label{mean-var}
\begin{aligned} 
& E[S^v] = \frac{1}{CT}\sum^{T-1}_{t=0} \sum^{C-1}_{k=0} S^{k,v}_t,
\\
& Var[S^v] =  \frac{1}{CT}\sum^{T-1}_{t=0}\sum^{C-1}_{k=0} (S^{k,v}_t - E[S^v])^2.
\end{aligned} 
\end{equation}

\begin{figure}[h]
\begin{center}
\includegraphics[width=12.5cm]{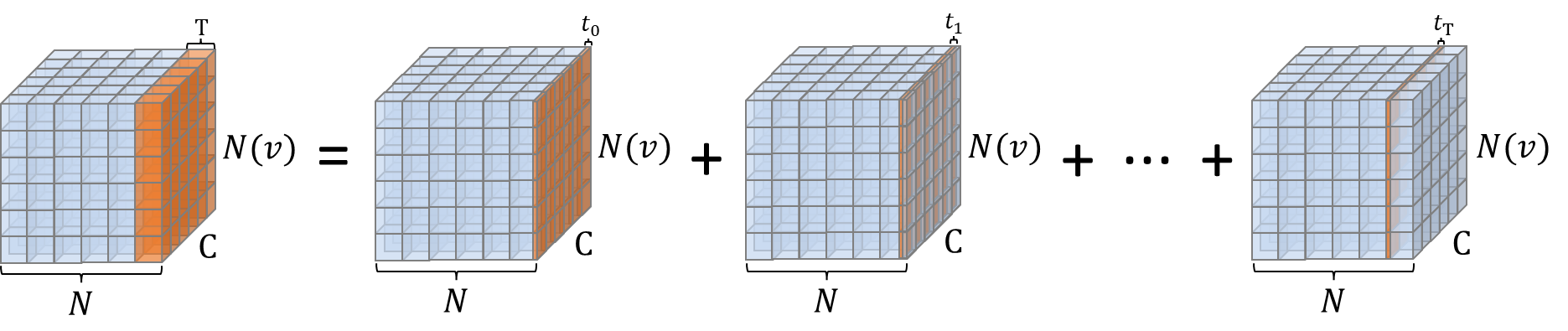}  %
\end{center}
\caption{The schematic diagram of spatial-temporal feature normalization (STFN), where the high-dimensional pre-synapse inputs will be normalized along the feature dimension and temporal dimension. 
}
\label{architecture}
\end{figure}
We follow the above schema and find the normalization can facilitate convergence, improve performance in training. 

\subsection{Graph SNNs Framework}
Our framework comprises two phases: spiking message passing and readout. The spiking message passing phase, detailed in Section 2.2, involves iteratively passing messages for \(T\) time steps during the inference process. After completing these iterations, the output spiking signals are decoded and transformed into high-level representations for downstream tasks. Under the rate coding condition, the readout phase involves computing the decoded feature vector for the entire graph using a readout function \(R\), as described by the following formula:
\begin{equation}
\label{readout}
\begin{aligned} 
\hat{y}^v = R(\{\frac{1}{T}\sum^{T-1}_{t=0}\widetilde{h}^v_t | v \in \mathcal{G} \}).
\end{aligned} 
\end{equation}
The readout function operates on the set of node states and must be invariant to permutations of these states to ensure that Graph SNNs are invariant to graph isomorphism. This means the function should produce the same output for isomorphic graphs, regardless of node order. The readout function can be implemented in various ways, such as using a differentiable function or a neural network that maps the node states to a final output, suitable for downstream tasks.

As depicted in Figure 1, the first phase of our framework involves two unfolding domains: the temporal domain and the spatial domain. In the temporal domain, the encoded binary node spikes are used as the current node feature at each time step \(t\). A sequence of binary information is processed sequentially, with historical information for each neuron playing a crucial role in spiking dynamics and output decoding.

In the spatial domain, the spiking features of a batch of nodes are first aggregated via edges connected to one-hop neighbors, using a propagation operation compatible with multiple proposed methods. These aggregated features are then fed into the spiking network module. Within each network module, we apply Spatial-Temporal Feature Normalization (STFN) to normalize the instantaneous membrane potential output along both the feature and temporal dimensions. This normalization ensures that the node features across all dimensions follow the distribution \(N(0,(\rho V_{th})^2)\). Each layer in the network encapsulates spatial-temporal dynamics for feature abstraction and mapping, and the stacked multi-layer structure enhances the SNNs' representation capabilities. The detailed inference procedure is provided in the Supplementary Materials.

To illustrate our approach, we use the semi-supervised learning task as an example. In this scenario, a small subset of node labels is available, and we evaluate the model's performance by calculating the cross-entropy error over all labeled examples. The formula for the cross-entropy loss in this context is as follows:
\begin{equation}
\label{loss}
\begin{aligned} 
\mathcal{L} = - \sum_{l\in \mathcal{Y}_L} \sum^{C^L-1}_{r=0} Y_{lr}ln \sigma(\hat{y}_{lr}), 
\end{aligned} 
\end{equation}
where \(\sigma(\cdot)\) denotes the softmax function, \(\mathcal{Y}_L\) represents the set of node indices with labels, and \(Y_{lr}\) denotes the true labels corresponding to the \(r_{th}\) dimension of the \(C^L\) classes. To effectively train the Graph SNN models using gradient descent, we employ the gradient substitution method in the backward pass\cite{wu2018spatio}. This involves using a rectangular function to approximate the derivative of the spike activity, allowing for the backpropagation of gradients through the spiking network.

\subsection{Coding Strategy}

In addition to the rate encoding scenario discussed previously, this framework also supports various other temporal encoding schemes. One such scheme is Rank Order Coding (ROC), which this paper explores to highlight the interesting properties of our graph learning framework under different encoding conditions. This demonstrates the framework's compatibility and scalability with various encoding methods.

Rank Order Coding assumes that biological neurons encode information based on the order of firing within a neuron ensemble. Consider a target neuron \(i\) receiving input from a presynaptic neuron set \(Q_n\) in the \(n\)-th layer, where each neuron fires only once, with its activation denoted as \(H^j\). ROC records the relative firing order of these neurons and updates the activation of the target neuron \(V^{n+1,i}\) as follows:
\begin{equation}
\label{roc}
\begin{aligned} 
V^{n+1,i}=\sum_{j \in Q_{n}} r^{order(H^{n,j})}{w_{i j}^{n+1}} 
\end{aligned}
\end{equation}
where $r \in (0,1)$ is a given penalty constant, and $\operatorname{order}\left(H^{n,j}\right)$ represents the firing order of neuron $j$ in the presynaptic ensemble. Equation (\ref{roc}) indicates that the sorting factor $r^{\text{order}\left(a_{j}^{l}\right)}$ is crucial for Rank Order Coding, encouraging neurons to fire early while penalizing those firing later. As this encoding scheme emphasizes information in the early spikes, it is well-suited for promoting network sparsity and facilitating rapid decision-making \cite{thorpe1998rank}. This encoding scheme can be seamlessly integrated into the propagation process of the graph learning framework, influencing membrane potential updates and spike generation. At the decision layer, a winner-takes-all strategy is employed, directly outputting the feature corresponding to the neuron with the fastest spike, enabling rapid decoding within short time steps. The characteristics of Rank Order Coding are advantageous for swift recognition and inference in static graph tasks.

\subsection{Model Instantiating: GC-SNN and GA-SNN}
To illustrate the effectiveness of our framework and normalization technique, we implement the framework into specific models by incorporating commonly used propagation methods in GNNs, such as graph convolution aggregators and graph attention mechanisms\cite{kipf2019semi, velivckovic2017graph}. In this implementation, our Graph Convolution Spiking Neural Network (GC-SNN) can be formulated as follows:
\begin{equation}
\widetilde{h}_t^{n,i} =  \Phi (\sum_{j\in \mathcal{N}(i)}\frac{1}{c_{ij}} \widetilde{h}_t^{n-1,j}W^n+ b^n, \widetilde{h}_{t-1}^{n}),
\label{gc-snn}
\end{equation}
where \(\mathcal{N}(i)\) represents the set of neighbors of node \(i\), and \(c_{ij}\) is the product of the square roots of the node degrees, specifically \(c_{ij} = \sqrt{|\mathcal{N}(i)|} \sqrt{|\mathcal{N}(j)|}\). The term \(b^n\) is a trainable bias parameter. Similarly, our Graph Attention Spiking Neural Network (GA-SNN) can be formulated as follows:
\begin{equation}
\begin{aligned} 
& \widetilde{h}_t^{n} =  \Phi (\alpha_{ij}^n \sum_{j \in \mathcal{N}(i)}\widetilde{h}_t^{n-1,j} W^n, \widetilde{h}_{t-1}^{n}),  \\ 
& \alpha_{ij}^n = \frac{exp(f_l({a^n}^T(\widetilde{h}_t^{n-1,i} W^n || \widetilde{h}_t^{n-1,j} W^n)))}{\sum_{k \in \mathcal{N}(i)}exp(f_l({a^n}^T(\widetilde{h}_t^{n-1,i} W^n || \widetilde{h}_t^{n-1,k} W^n)))},  \\
\end{aligned} 
\label{gc-snn}
\end{equation}
where \(f_l(\cdot)\) denotes the LeakyReLU activation function, \(LeakyReLU(\cdot)\). The vector \(a^n\) is a learnable weight vector, and \(\|\| \) represents the concatenation operation. The core idea is to apply an attention mechanism to graph convolution, allowing the model to learn different importance weights, \(\alpha\), for each node. The experimental section presents the results of the models discussed above for further illustration.

\subsection{Spatial-temporal Embedding for Downstream Tasks}

We further specify node and graph classification tasks on multi-graph datasets to validate the  effectiveness of our method. For different downstream tasks, we generate different level representations by using the output of the last GCN layer and provide a loss function for training the network parameters.

\textbf{Node classification task.} Since the node features can be obtained directly from the GCN output, we directly input the node features into the MLP to obtain logits $y_{i, \text { pred }} \in \mathbb{R}^{C}$  for each class\cite{dwivedi2020benchmarking}. The formula can be expressed as:\\
\begin{equation}
y_{i, \text { pred }}=A\operatorname{ReLU}\left(W \widetilde{h}_t^{n,i}\right)
\end{equation}
\text {where } $A \in \mathbb{R}^{d \times C}$, $W \in \mathbb{R}^{d \times d}$, and the $\widetilde{h}_t^{n,i}$ represents the $i^{th}$ node's feature of the last GCN layer $n$ at the last time step $t$  \text {. }To train the network parameters, we use cross entropy between logits and the ground truth labels as the loss function.


\textbf{Graph classification task.} For graph classification tasks, we take the average of node features outputted by GCN to generate a representation of the entire graph $y_{\mathcal{G}}$. The formula can be expressed as:
\begin{equation}
y_{\mathcal{G}}=\frac{1}{\mathcal{V}} \sum_{i=0}^{\mathcal{V}} \widetilde{h}_t^{n,i}
\end{equation}
where $\widetilde{h}_t^{n,i}$ represents the $i^{th}$ node's feature of the last GCN layer $n$ at the last time step $t$ . Then the representation of the graph is inputted into an MLP to obtain logits $y_{pred}$\cite{dwivedi2020benchmarking}. 
\begin{equation}
y_{pred}=A\operatorname{ReLU}\left(W y_{\mathcal{G}}\right)
\end{equation}
\text {where } $A \in \mathbb{R}^{d \times C}$, $W \in \mathbb{R}^{d \times d}$ \text {. }Afterward, the cross-entropy between the logits and ground truth labels is used as the error for training.

\label{dataset}

\section{Experiments}
In this section, We firstly investigate the capability of our models over semi-supervised node classification on three citation datasets to examine their performance. Then we demonstrate the effectiveness of the proposed STFN and provide some analysis visualization for the model efficiency. 

\begin{table}[h]
\centering
\begin{tabular}{lcccc}
\toprule
Dataset  &  Graphs  &  Avg. Nodes  & Avg. Edges & Task Type \\
\midrule
Cora & 1 & 2708 & 5429 & Node Classification \\
Pubmed & 1 & 19717 & 44338 & Node Classification \\
Citeseer & 1 & 3327 & 4732 & Node Classification \\
Pattern & 14000 & 117.47 & 4749.15 & Node Classification \\
Cluster & 12000 & 117.20 & 4301.72 & Node Classification \\
MNIST & 70000 & 70.57 & 564.53 & Graph Classification \\
CIFAR10 & 60000 & 117.63 & 941.07 & Graph Classification \\
\bottomrule
\end{tabular}
\caption{Summary of basic information for datasets used in this work. }
\label{dataset}
\end{table}

\begin{table}[h]
\centering
\resizebox{\textwidth}{!}{%
\begin{tabular}{lcccc}
\toprule
Dataset & Classes &  $\sum \text{Nodes}$ & Node feat. & Training/Validation/Testing \\
\midrule
Cora & 7 & 2708 & Word Vector(1433) & 140/500/1000(Nodes) \\
Pubmed & 3 & 19717 & Word Vector(500) & 60/500/1000(Nodes) \\
Citeseer & 6 & 3327 & Word Vector(3703) & 120/500/1000(Nodes) \\
Pattern & 2 & 1664491 & Node Attr(3) & 10000/2000/2000(Graphs) \\
Cluster & 6 & 1406436 & Node Attr(7) & 10000/1000/1000(Graphs) \\
TSP & 2 & 3309140 & Coord(2) & 10000/1000/1000(Graphs) \\
MNIST & 10 &  4939668 & Pixel+Coord(3) & 55000/5000/10000(Graphs) \\
CIFAR10 & 10 & 7058005 & Pixel(RGB)+Coord(5) & 45000/5000/10000(Graphs) \\
\bottomrule
\end{tabular}}
\caption{Overview of the datasets employed in this study.}
\label{dataset}
\end{table}

\subsection{Datasets and Pre-processing}
\textbf{Basic experiments.} We first use three standard citation network benchmark datasets---Cora, Pubmed, and Citeseer, where nodes represent paper documents and edges are (undirected) citation links.  We summarize the dataset statistics used in our experiments in Table \ref{dataset}. The datasets contain sparse bag-of-words feature vectors for each document and a list of citation links between documents. In Table \ref{dataset}, nodes represent paper documents and edges represent citation links. Label rate denotes the number of labels used for training, and features denote the dimension of feature vector for each node. The Cora dataset contains 2708 nodes, 5429 edges, 7 classes and 1433 features per node. The Pubmed dataset contains 19717 nodes, 44338 edges, 3 classes and 500 features per node. The Citeseer dataset contains 3327 nodes, 4732 edges, 6 classes and 3703 features per node. Each document node has a class label. We only use 20 labels per class during training with all feature vectors.

We model the citation links as (undirected) edges and construct a binary, symmetric adjacency matrix $A$. Note that node features correspond to elements of binary bag-of-words representation. Thus, We treat the binary representations as spike vectors, and re-organize them as an assembling sequence set w.r.t timing, where each component vector is considered equal at different time steps for simplicity. In this manner, we train our SNN modes with spike signals and evaluate them on 1000 test nodes, and we use 500 additional nodes for validation following the configuration used in the work \cite{kipf2019semi}. 

We reproduce GCN and GAT and implement our GC-SNN and GA-SNN models with deep graph library. The experimental settings of models with same propagation operator (graph convolution and graph attention) are kept the same in each dataset for fairness. We use Adam optimizer \cite{kingma2014adam} with an initial learning rate of $0.01$ for GCN and GC-SNN, and $0.005$ for GAT and GA-SNN. We use the dropout technique for avoiding over-fitting, which is set to $0.1$ for GC-SNN and $0.6$ for GA-SNN model. All models run for $200$ epochs and are repeated $10$ trials with different random seeds. In each trial, the models are initialized by a uniform initialization and trained by minimizing the cross-entropy loss on the training nodes. We use an L2 regularization with weight decay as $0.0005$, and only use the available labels in training set and test the accuracy using $1000$ testing samples. 

For Graph SNN, we set the time window $T$ as $8$, and set the threshold $V_{th}$ as $0.25$ for basic performance evaluation. We use the Integrate-and-Fire (IF) neurons in our models, where the historical membrane potential will not decay as time step continues. Specially, in order to train the Graph SNN models effectively via gradient descent paradigm, we adopt gradient substitution method in backward path \cite{wu2018spatio} and take the rectangular function to approximate the derivative of spike activity. It yields

\begin{equation}
r(u) = \frac{1}{\nu}sign(|V-V_{th}|<\frac{\nu}{2},
\end{equation}
where $\nu$ represents the width of $r(u)$, and set as $0.5$ in all experiments. 

For GAT and GA-SNN, we adopt an MLP structure [Input-64-Output] with $8$ attention heads. For GCN and GC-SNN, we adopt an MLP structure [Input-400-16-Output]. In view of the linear propagation operators used in this work, we follow the formula $(8)$ and conduct the feature transformation first then perform aggregation operation. Besides, we use the proposed STFN technique for GA-SNN and GC-SNN. All experiments in this work are implemented by PyTorch with an acceleration of 4 RTX 2080Ti GPUs.

\textbf{Extended experiments.} In addition, we conduct extended experiments for verifying the scability of our model on multi-graph datasets regarding node and graph level. 

\textbf{Node level.} Further, we used two datasets, Pattern and Cluster, for the node classification task on multiple graphs, which are generated with the Stochastic Block Model (SBM). SBM is a traditional graph generation model in which each node belongs to a different community, and each community is connected with different probabilities\cite{abbe2017community}. 

The Pattern dataset is used to identify a specific graph pattern $P$ within larger graphs $G$ of varying sizes. In this dataset, graphs $G$ were generated with $5$ communities of random sizes between $5$ and $35$. The intra-communities and extra-communities probabilities for each community were set at $0.5$ and $0.35$ respectively, and node features were generated using a uniform random distribution with a vocabulary of size $3$, i.e., $\{0, 1, 2\}$. The Pattern dataset contains 14,000 graphs, 1,664,491 nodes, 66,488,100 edges, and 2 classes. And it includes 10,000 graphs for training, 2,000 for validation, and 2,000 for testing.

The Cluster dataset is used for semi-supervised clustering. It includes $6$ clusters which are generated by the Stochastic Block Model (SBM) with random sizes between $5$ and $35$, and intra-communities and extra-communities probabilities of $0.55$ and $0.25$, respectively. Each node is given an input feature value chosen from the set $\{0, 1, 2, ..., 6\}$. If the value is $1$, it is assigned to class $0$, if the value is $2$, it is assigned to class $1$, and so on till value $6$, which is assigned to class $5$. If the value is $0$, then the class of the node is undetermined and will be determined by the GNN. One labeled node is randomly assigned to each community, and most node features are set to $0$. The Cluster dataset contains 12,000 graphs, 1,406,436 nodes, 51,620,640 edges, and 6 classes. And it includes 10,000 for training, 1,000 for validation, and 1,000 for testing.



\textbf{Graph level.} In the graph level, we convert each image in the popular MNIST and CIFAR10 datasets into graph using super-pixels\cite{knyazev2019understanding} and classify these graphs. The node features of the graph are generated by the intensity and position of the super-pixels, and the edges are $k$ nearest neighbor super-pixels. For both MNIST and CIFAR10, the value of $k$ is set to $8$. The resulting graphs are of sizes 40-75 nodes for MNIST and 85-150 nodes for CIFAR10.

The MNIST dataset contains 70000 graphs, 4939668 nodes, 39517100 edges, 10 classes and it includes 55,000 graphs for training, 5,000 for validation, and 10,000 for testing. The CIFAR10 dataset contains 60,000 graphs, 7,058,005 nodes, 56,464,200 edges, 10 classes and it includes 45,000 graphs for training, 5,000 for validation, and 10,000 for testing.

\begin{table}[H]
\centering
\resizebox{\textwidth}{!}{%
\begin{tabular}{lcc}
\toprule
Dataset & GCN Network Structure & GC-SNN Network Structure \\
\hline Pattern & Input-GCNLayer146*4-MLP-Output & Input-GCSNNLayer146*4-MLP-Output \\
Cluster & Input-GCNLayer146*4-MLP-Output & Input-GCSNNLayer146*4-MLP-Output \\
MNIST & Input-GCNLayer146*4-MLP-Output & Input-GCSNNLayer146*4-MLP-Output \\
CIFAR10 & Input-GCNLayer146*4-MLP-Output & Input-GCSNNLayer146*4-MLP-Output \\
\bottomrule
\end{tabular}}
\caption{
Network structure configuration.}
\label{performance}
\end{table}

\begin{table}[H]
\centering
\resizebox{\textwidth}{!}{%
\begin{tabular}{lcccccc}
\toprule
Model & Hyper-parameter & Pattern & Cluster & MNIST & CIFAR10 \\
\hline \multirow{4}{*}{\text { Shared }} & \text { Max Epoch } & 1000 & 1000 & 1000 & 1000 \\
& \text { Batch Size } & 128 & 128 & 128 & 128 \\
& \text { Initial Learning Rate } & 0.001 & 0.001 & 0.001 & 0.001 \\
& \text { Minimum Learning Rate } & $1e^{-5}$ & $1e^{-5}$ & $1e^{-5}$ & $1e^{-5}$ \\
\hline \multirow{4}{*}{\text { SNNs }} & \text { Firing Threshold } & 0.25 & 0.25 & 0.25 & 0.25 \\
& \text { Leakage Factor } & 1.0 & 1.0 & 1.0 & 1.0 \\
& \text { Gradient Width } & 0.5 & 0.5 & 0.5 & 0.5 \\
& \text { Time Window } & 8 & 8 & 8 & 8 \\
\bottomrule
\end{tabular}}
\caption{
Hyper-parameter setting for extended experiments.}
\label{performance}
\end{table}

We reproduce the GCN and implement our GC-SNN models using PyTorch. To ensure fairness, as listed in Table 3 and 4, the experimental settings for models with the same graph convolution operator were kept the same for each dataset. We used the Adam optimizer with an initial learning rate of $0.01$ for both GCN and GC-SNN. We employed the dropout technique to prevent overfitting, which was set to $0.1$ for the GC-SNN model. We use cross-entropy as the loss function on the training nodes.

For the GC-SNN, we set the time window $T$ as $15$ and the threshold $V_{th}$ as $0.25$ for basic performance evaluation. We used Integrate-and-Fire (IF) neurons in our models, where the historical membrane potential does not decay as the time step continues. To train the Graph SNN models effectively via gradient descent, we adopt the gradient substitution method in the backward path and use the rectangular function to approximate the derivative of spike activity.


\subsection{Performance Verification}

\begin{table}
\centering
\resizebox{\textwidth}{!}{%
\begin{tabular}{l|ccccccc|cc}
\toprule
Dataset  &  SemiEmb  &  DeepWalk  & ICA & Planetoid & Chebyshev & GCN$^\ast$ & GAT$^\ast$ & GC-SNN(Ours) & GA-SNN(Ours)\\
\hline Cora & 59.0 & 67.2 & 75.1 & 75.7 & 81.2 & 81.9$\pm$1.1 & 82.3$\pm$0.6 & \textbf{80.7}$\pm$0.6 & 79.7$\pm$0.6 \\
Pubmed & 71.7 & 65.3 & 73.9 & 77.2 & 74.4 & 79.4$\pm$0.4 & 78.4$\pm$0.5 & \textbf{78.1}$\pm$0.3 & 78.0$\pm$0.4 \\
Citeseer & 59.6 & 43.2 & 69.1 & 64.7 & 69.8 & 70.4$\pm$1.1 & 71.1$\pm$0.2 & \textbf{69.9}$\pm$0.9 & 69.1$\pm$0.5 \\
\bottomrule
\end{tabular}}
\caption{
Performance comparison on benchmark datasets \protect\cite{perozzi2014deepwalk,getoor2005link,yang2016revisiting,defferrard2016convolutional}. $\ast$ denotes the results in our implementation and $\pm$ denotes the standard deviation calculated from $10$ runs.}
\label{performance}
\end{table}



\paragraph{Basic Performance} For fairness, we take the same settings for GNN models (GCN, GAT) and our SNN models (GC-SNN, GA-SNN), and also keep the settings of each dataset same. We report our results for 10 trials in Table $\ref{performance}$. The results suggest that even if using binary spiking communication, our SNN models can achieve comparable performance with the SOTA results with a minor gap. It proves the feasibility and powerful capability of spiking mechanism and spatial-temporal dynamics, which can bind diversiform features from different nodes and work well on graph scenarios with few labels. 

\begin{figure*}[h]
\begin{center}
\includegraphics[width=12.0cm]{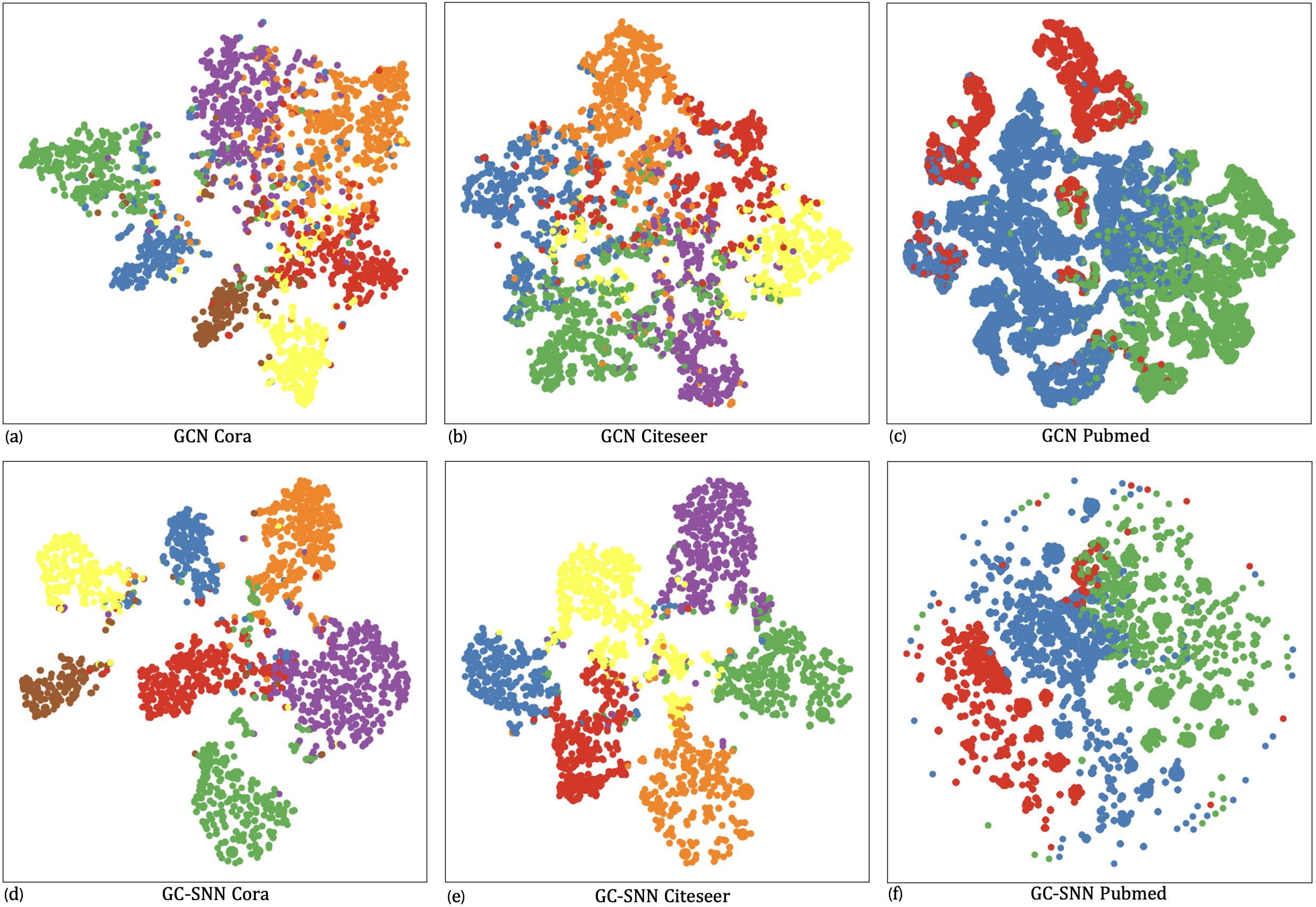}
\end{center}
\caption{(a)(b)(c)Visualization of the last hidden layer features of the GCN on the Cora, Citeseer, and Pubmed datasets produced by t-SNE. (d)(e)(f)Visualization of the last hidden layer features of the GC-SNN on the Cora, Citeseer, and Pubmed datasets produced by t-SNE. 
}
\label{tsne}
\end{figure*}

To more intuitively demonstrate the advantage of GC-SNN model compared with GCN model, we use t-SNE to visualize the final hidden layer features down to 2 dimensions. t-SNE is a nonlinear dimensionality reduction algorithm that can well reduce high-dimensional data to 2 dimensions for visualization. As shown in Figure $\ref{tsne}$, when the features obtained with the GC-SNN hidden layers, the samples within the same class are more concentrated and the inter-class spacing is larger and more separated.
Specifically, on the Cora, Citeseer, and Pubmed datasets, our SNN model demonstrated superior performance in distinguishing between different classes, thereby exhibiting stronger feature representation capabilities compared to GCN.

\paragraph{Extended Performance} 
When conducting the extended dataset for node-level and graph-level tasks on multi-graph datasets, we used the same settings for the same models to ensure the fairness of the experiments. As shown in Table 6, our model demonstrates similar performance to other existing models, proving that GC-SNN can be adapted to different tasks with good performance.


\begin{table}[H]
\centering
\resizebox{\textwidth}{!}{%
\begin{tabular}{lccccc}
\toprule
Dataset & MLP & GCN & GAT & GIN & GC-SNN \\
\midrule
Pattern(Acc. $\pm$ s.d.)   &  50.52$\pm$0.00  & 85.50$\pm$0.05  & 75.82$\pm$1.82 & 85.59$\pm$0.01 & 85.43$\pm$0.13  \\
Cluster(Acc. $\pm$ s.d.)   &  20.97$\pm$0.00  & 47.83$\pm$1.51  & 57.73$\pm$0.32 & 58.38$\pm$0.24 & 53.06$\pm$1.33 \\
\midrule
MNIST(Acc. $\pm$ s.d.)     &  95.34$\pm$0.14  & 90.12$\pm$0.15  & 95.54$\pm$0.21 & 96.49$\pm$0.25 & 92.76$\pm$0.00   \\
CIFAR10(Acc. $\pm$ s.d.)   &  56.34$\pm$0.18  & 54.14$\pm$0.39  & 64.22$\pm$0.45 & 55.26$\pm$1.53 & 51.75$\pm$0.01  \\
\bottomrule
\end{tabular}}
\caption{
Performance comparison on extended datasets \protect\cite{perozzi2014deepwalk,getoor2005link,yang2016revisiting,defferrard2016convolutional}. $\ast$ denotes the results in our implementation and $\pm$ denotes the standard deviation calculated from $10$ runs.}
\label{performance}
\end{table}

\subsection{Ablation Study for the Impact of STFN}

\begin{figure*}[h]
\begin{center}
\includegraphics[width=12.0cm]{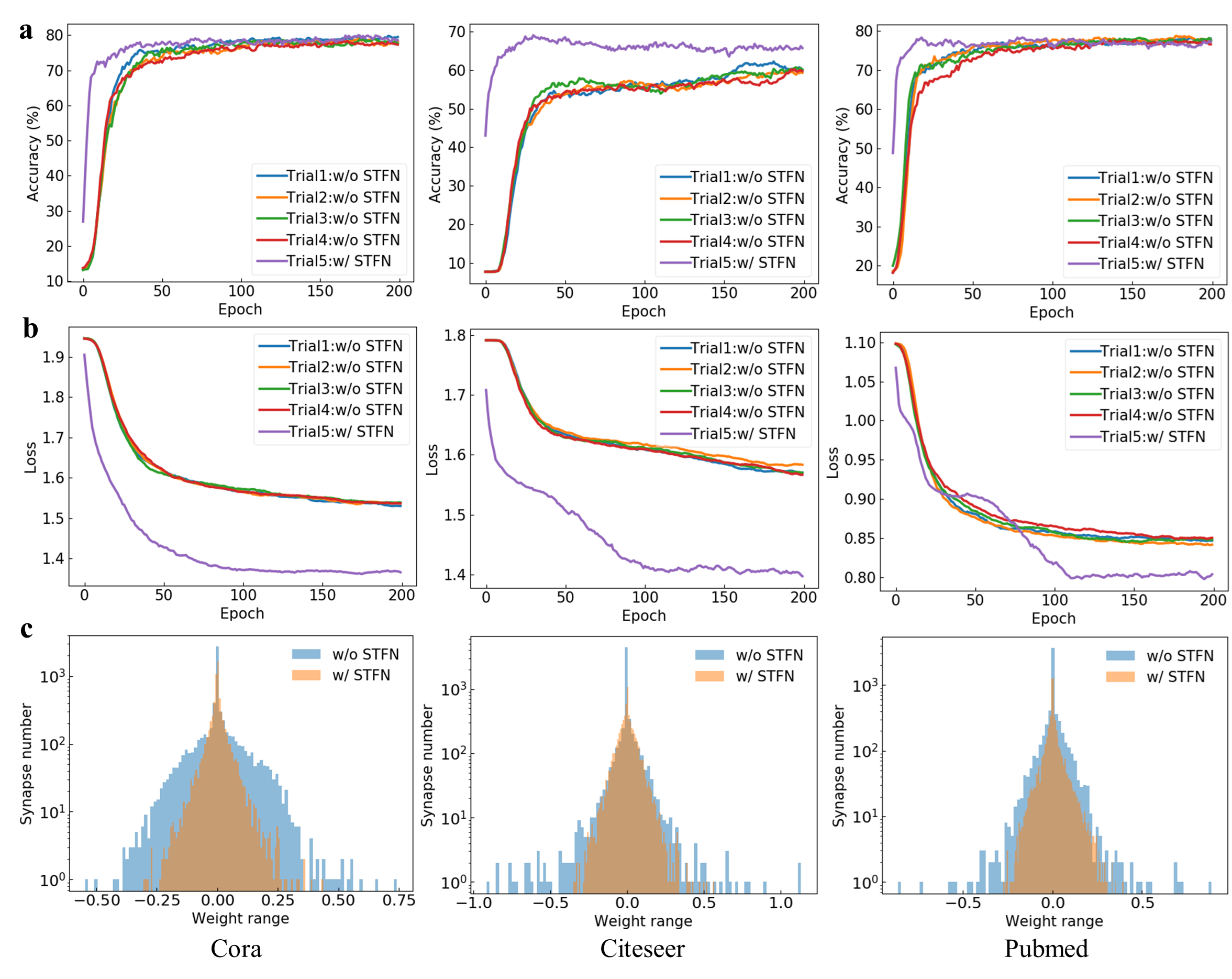}   
\end{center}
\caption{(a) Validation accuracy variations during the training of Graph SNN across Multiple datasets. (b) Loss variations during the training of Graph SNN across Multiple datasets. (c) Comparative visualization of weight distributions learned by Graph SNN across multiple datasets. 
}
\label{abla}
\end{figure*}

To further validate the impact of STFN on SNNs in graph learning tasks, this study conducted systematic ablation experiments on the Cora, Citeseer, and Pubmed datasets. Figures $\ref{abla}(a)(b)$ depict the validation accuracy curves and loss convergence curves during the training process, respectively. The results reveal that STFN exhibits advantages in two main aspects: firstly, it enhances the generalization capability of SNNs, leading to higher recognition accuracy on test samples across all datasets; secondly, it significantly accelerates convergence. Compared to the baseline SNN model without STFN, the SNN with STFN achieves faster convergence on all three datasets, with substantially lower loss errors. These observations demonstrate that STFN facilitates faster convergence, alleviates overfitting, and significantly enhances performance in semi-supervised learning tasks for SNNs.

To investigate the influence of STFN on learned synaptic weights, we further visualizes the weight data distribution in Figure $\ref{abla}(c)$. The results indicate that STFN can concentrate the distribution of synaptic weights around the zero-value range. In this scenario, the normalized membrane potential state can coordinate distribution differences and pulse-triggering thresholds, resulting in sparser weights. This is advantageous for the deployment of neuromorphic hardware and the development of low-power graph computing applications.


\subsection{Rate Coding versus Rank Order Coding}

\begin{figure*}[h]
\begin{center}
\includegraphics[width=12.2cm]{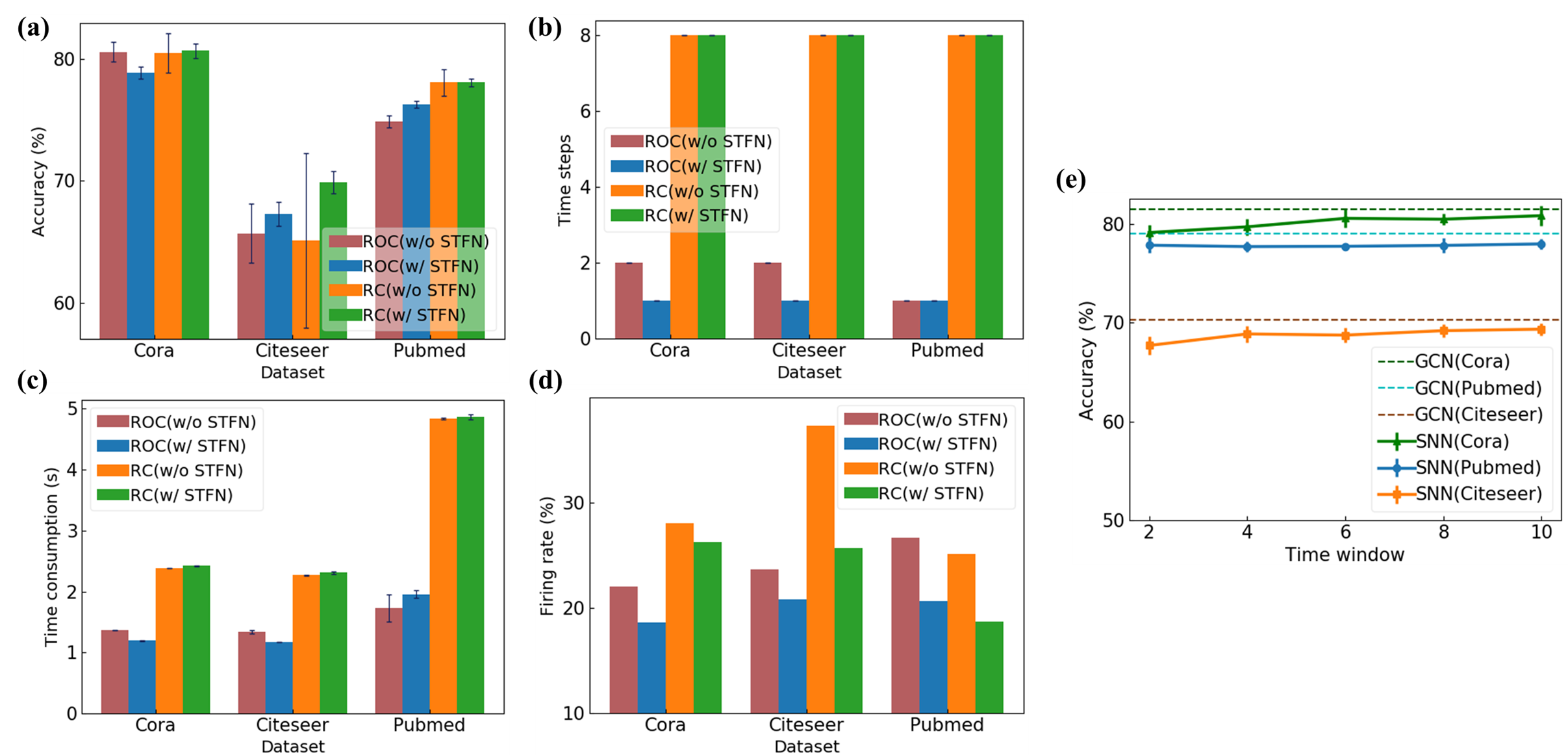}   
\end{center}
\caption{(a) Accuracy comparison; (b) Inference time steps comparison; (c) Time cost per training epoch; (d) Firing rate comparison; (e) Impact of time window length on rate encoding.
}
\label{rank-rate}
\end{figure*}

To investigate the performance differences under different encoding schemes, this study conducted comparative experiments between rate coding (RC) and rank-order coding (ROC). As shown in Figure \ref{rank-rate}, the performance gap of the Graph SNN under the two encodings is relatively small. For graph learning tasks, ROC achieves slightly lower accuracy than RC, but with significantly reduced inference time steps compared to RC. STFN further promotes the recognition accuracy under both encodings in the Citeseer and Pubmed datasets, while also reducing the inference time steps and training time cost for the ROC network (see Figure \ref{rank-rate}(b)(c)). This indicates that temporal encoding, at the expense of slight accuracy loss, enhances network inference speed and learning efficiency. Additionally, through firing rate analysis, this study found that ROC results in a lower spike firing rate for the network (see Figure \ref{rank-rate}(d)). This phenomenon suggests that temporal encoding can reduce the neural activity intensity in the network, consistent with the sparse neural connections shown in Figure \ref{rank-rate}(c). A more sparse weight distribution and neural firing will facilitate low-power computation and hardware deployment for the network.


Moreover, this study also investigates the impact of the time window length on the RC network. As depicted in Figure $\ref{rank-rate}(e)$ consistently across the three datasets, a longer time window length contributes to a gradual improvement in accuracy, while the dependency of RC on the time window length is not very pronounced. Clearly, in scenarios where high precision is not a strict requirement, a smaller time window can be used to reduce training costs, and the model can still achieve satisfactory results. In situations where efficiency and speed are paramount, the adoption of the ROC strategy can be beneficial to reduce latency and neural activity.

\subsection{Investigation into Oversmoothing Issue in Deeper Structure}
The oversmoothing problem refers to the phenomenon in graph representation learning where the application of graph-based learning algorithms, such as GNNs, results in excessive smoothing of node representations, leading to a reduction in dissimilarity between nodes. Specifically, the oversmoothing problem occurs in models like GNNs, which learn node representation vectors by propagating and aggregating information between nodes. During this process, information from neighboring nodes is aggregated and updated into the representation of the target node. After multiple iterations, the node representations tend to average out, making the representations of different nodes become similar and diminishing the distinguishability between nodes. This implies that the model's performance degrades with increasing depth of layers. It is generally believed that the problem is primarily caused by the aggregation operations and multiple iteration updates during the information propagation process, leading to information loss and blurring. When information from neighboring nodes is excessively aggregated, the representation vectors of nodes tend to become similar, failing to accurately reflect the individual characteristics and distinctions of nodes. Although some research has attempted to alleviate this problem by adjusting aggregation operations or the graph model, the oversmoothing problem remains a significant and unresolved challenge in graph representation learning.

\begin{figure*}[h]
\begin{center}
\includegraphics[width=12.0cm]{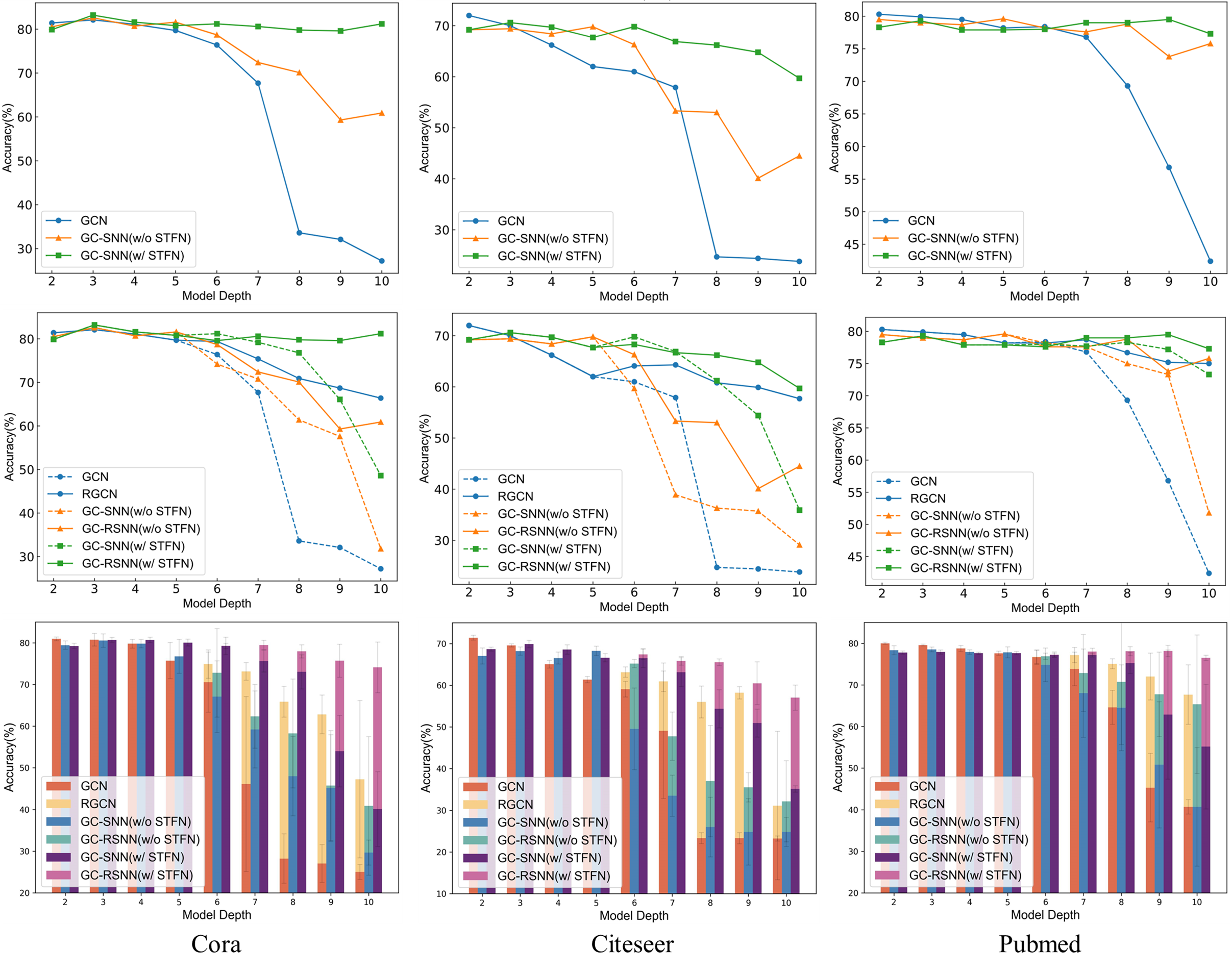}   
\end{center}
\caption{(a) Ablation experiments on the impact of STFN on network performance degradation; (b) Ablation experiments on the impact of residual structure and STFN on network performance degradation; (c) Performance comparison of models with different depths and the performance improvement gains brought by residual structure or STFN.
}
\label{oversmooth}
\end{figure*}

Based on this, through extensive experiments, this paper discovered a significant performance degradation caused by the oversmoothing problem in the graph spiking neural network model. As shown in Figure \ref{oversmooth}(a), the GCN model represented by the blue line experiences severe performance degradation. The instantiated graph convolution spiking neural network, GC-SNN (w/o STFN), without applying spatiotemporal normalization, also exhibits noticeable performance degradation (see the orange curve in Figure \ref{oversmooth}(a)), with a decreasing trend in classification accuracy as the network depth increases. However, overall, compared to GCN, GC-SNN achieves higher accuracy after the decline, especially as the depth increases, and this difference becomes more significant. For example, on the Pubmed dataset, a ten-layer GC-SNN outperforms GCN by around 35$\%$, indicating that the spiking graph learning framework constructed in this paper, along with the characteristics of spiking dynamics, is effective and helps alleviate the oversmoothing problem in graph representation.

To further investigate the impact of STFN on this problem, this paper compares the performance of Graph SNN models with and without STFN (see the green curve in Figure \ref{oversmooth}(a)). Clearly, based on the same residual network structure, the model's performance degradation problem is largely eliminated by applying STFN, except for a slight decrease on the Citeseer dataset. On the other two datasets, the performance shows a smooth upward or oscillating trend. This phenomenon indicates that STFN technology can indeed further enhance the graph learning capability of spiking networks, effectively overcoming the performance degradation caused by deep oversmoothing.

\begin{figure*}[h]
\begin{center}
\includegraphics[width=12.0cm]{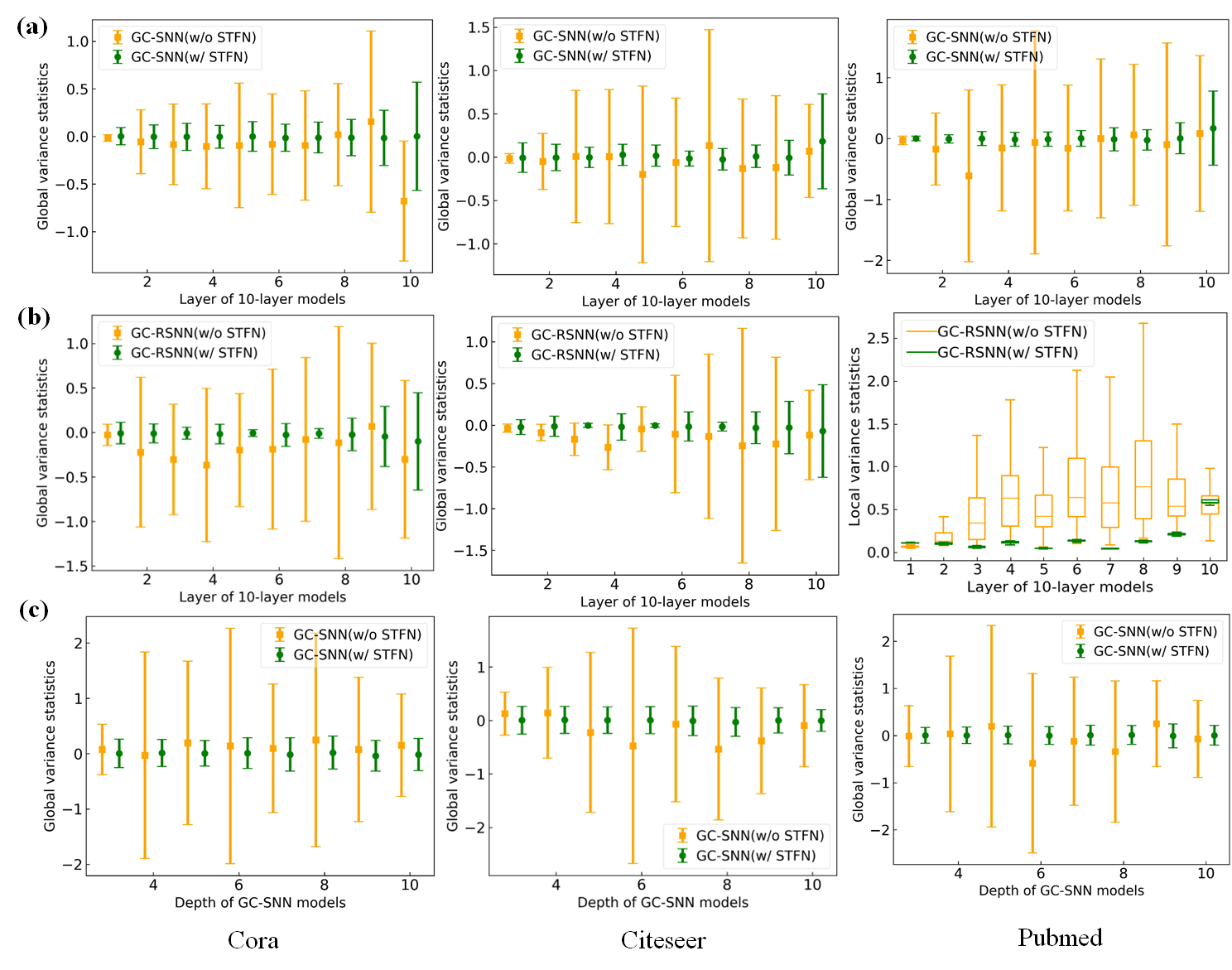}   
\end{center}
\caption{(a) Global variance statistics of features for each hidden layer in the pre-trained ten-layer GC-SNN model; (b) Global variance statistics of features for each hidden layer in the pre-trained ten-layer GC-SNN model with residual structure; (c) Training GC-SNN models with different depths and then performing global variance statistics on the features of their final hidden layer.
}
\label{global-var}
\end{figure*}

Considering the performance degradation in deep networks may be caused by the vanishing gradient problem and to eliminate the interference of the residual network structure in overcoming the performance degradation, this paper further conducts ablation experiments to explore the impact of this factor. Figure \ref{oversmooth}(b) presents the test results when the residual structure is ablated on the basis of the original network model, where RSNN represents ResNet-based SNN. Different from traditional ResNet, a custom-designed two-layer small residual block is used here as the connection and computation unit. It can be seen that without STFN and only the residual structure, GC-RSNN achieved limited improvement compared to GC-SNN (see the orange dashed and solid lines in Figure \ref{oversmooth}(b)), but overall still showed a downward trend, indicating that the residual prior can alleviate the performance decrease to some extent due to the vanishing gradient but does not address the fundamental issue of oversmoothing. Similarly, this holds true for GCN as well (see the blue dashed and solid lines in Figure \ref{oversmooth}(b)). Figure \ref{oversmooth}(c) displays the statistical contribution of STFN and the residual structure to performance improvement in multiple experiments. Clearly, STFN plays a major role in enhancing the model's performance, further demonstrating the efficiency and potential of the proposed pulse graph learning framework and the spatiotemporal feature normalization algorithm.

\begin{figure*}[h]
\begin{center}
\includegraphics[width=12.0cm]{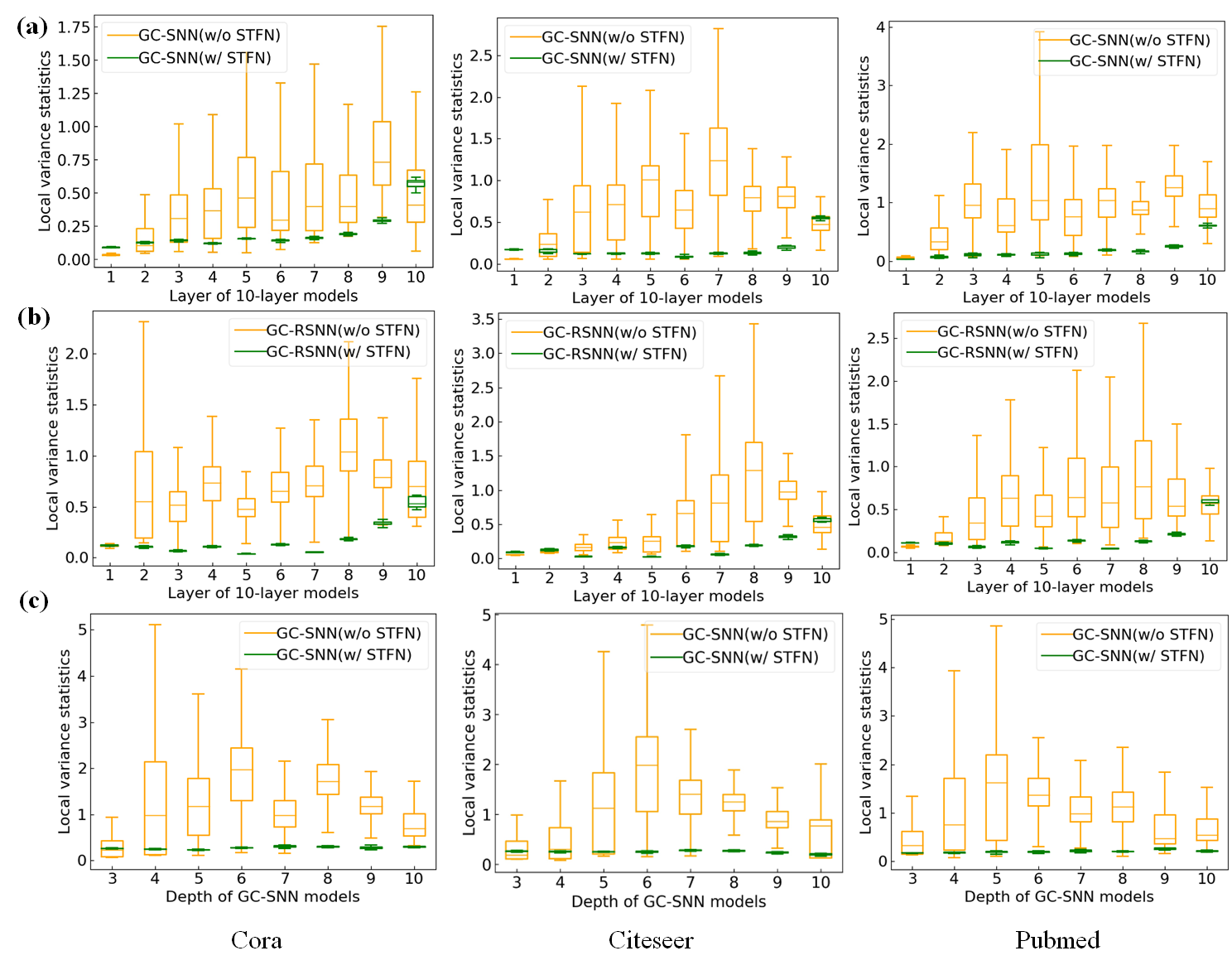}   
\end{center}
\caption{(a) Local variance statistics of features for each hidden layer in the pre-trained ten-layer GC-SNN model; (b) Local variance statistics of features for each hidden layer in the pre-trained ten-layer GC-SNN model with a residual structure; (c) Local variance statistics of features for the final hidden layer of GC-SNN models trained at different depths.
}
\label{local-var0}
\end{figure*}

On the other hand, this paper investigates the underlying reasons for the improvement brought by STFN from the perspective of representation. Figure \ref{global-var} presents a statistical analysis of differences in hidden layer representations at different layers within different networks. Global variance characterizes the overall diversity of features across all nodes in a graph. It is calculated by combining the feature vectors of all nodes and computing the mean and variance for the entire set. As shown in Figure \ref{global-var}(a)(b), regardless of the presence of a residual structure (GC-SNN or GC-RSNN), models with STFN have lower global variance compared to models without STFN, and their mean points stabilize around zero. This indicates that STFN can make the features learned by each layer of SNN more stable globally, with overall feature differences across the entire graph being smaller. Similarly, for SNN models of different depths, this paper further calculates the global variance of the final hidden layer, as shown in Figure \ref{global-var}(c). The results indicate that, for different depths of GC-SNN models, models with STFN exhibit stronger stability and smaller fluctuations in the features of the internal hidden layers they learn.

In addition, this study further calculates the feature differences between nodes, referred to as local variance, where the feature vectors of nodes are considered, and the statistical variance among nodes within the entire graph is computed. The results, as shown in Figure \ref{local-var0}, indicate that STFN also reduces the local variance of node features in GC-SNN. This reflects a consistent stability in local differences, similar to the global differences. Therefore, it is inferred that STFN enables the SNN to overcome the performance degradation caused by oversmoothing by promoting a more consistent representation of hidden layers within the network. Related works in graph representation learning suggest that excessively large variances in node features are fundamental reasons for performance degradation \cite{zhou2021understanding}. Further analysis of STFN reveals that normalization essentially adjusts the node variance. Assuming the denominator of the STFN normalization is in a generic form, denoted as $\frac{1}{p}$, with a mean given by
$\hat{x}_k^t = \frac{{\lambda V_{th} (x_k^t - E[x_k^t])}}{{(\sigma[x_k^t] + \epsilon)^{\frac{1}{p}}}}$, the normalized variance is then:
\begin{figure*}[h]
\begin{center}
\includegraphics[width=12.0cm]{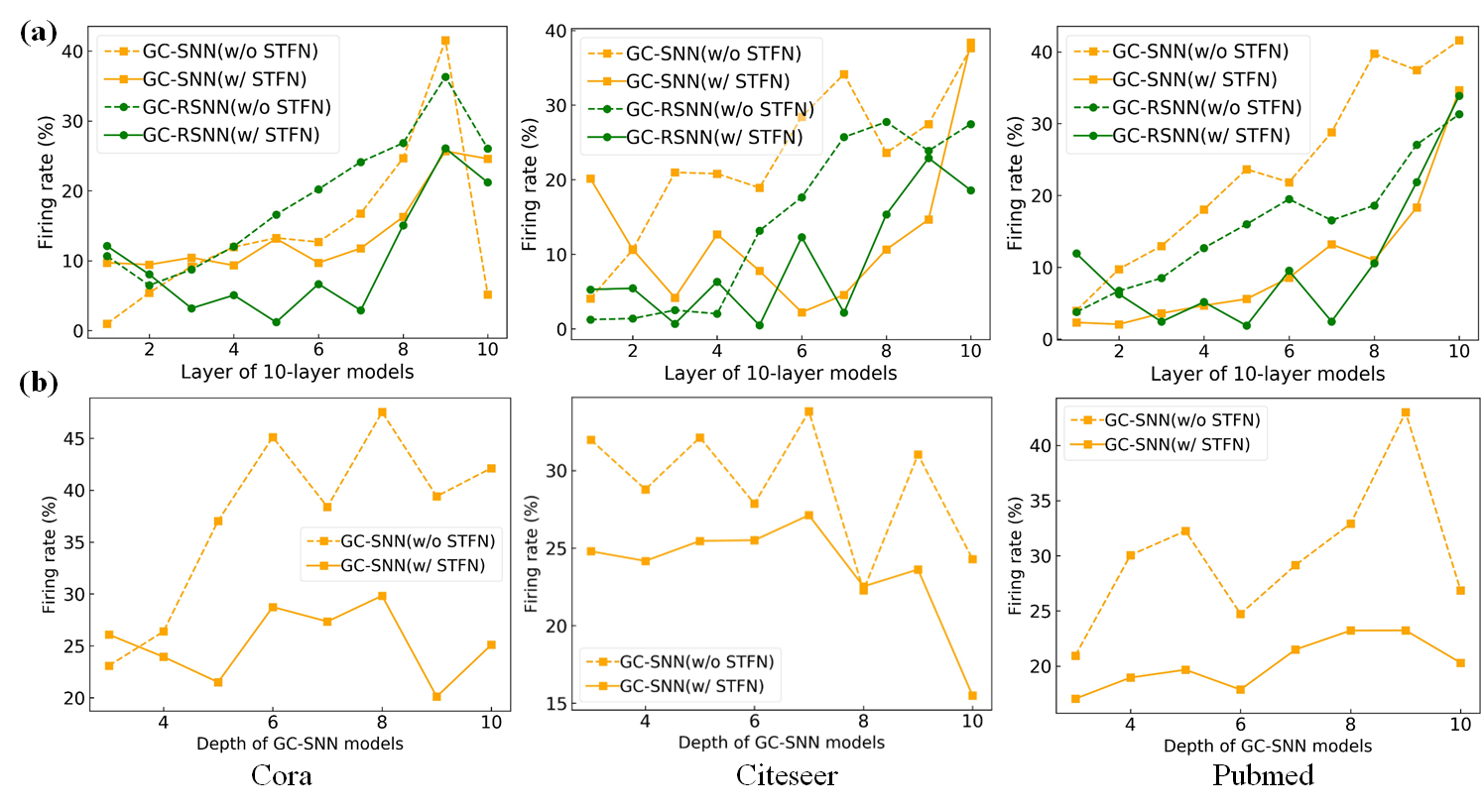}   
\end{center}
\caption{(a) Average firing rates of pre-trained ten-layer GC-SNN and GC-RSNN models at different layers; (b) Statistical analysis of the average firing rates of the final hidden layer for trained GC-SNN models of different depths.
}
\label{fr}
\end{figure*}
\begin{equation}
\label{stfn-var}
\begin{split}
    \hat{\sigma} &= \sqrt{\frac{1}{K} \sum_{j=1}^{K} \left(\hat{x}_k^t - \mathbb{E}\left[x_k^t\right]\right)^2} \\
    &= \sqrt{\frac{1}{K} \sum_{j=1}^{K} \left(\frac{x_k^t - \mathbb{E}\left[x_k^t\right]}{\sigma^{1/p}} - \frac{\mathbb{E}\left[x_k^t\right] - \mathbb{E}\left[x_k^t\right]}{\sigma^{1/p}}\right)^2} = \sigma^{(1-\frac{1}{p})}
\end{split}
\end{equation}
Equation (\ref{stfn-var}) demonstrates the controlling effect of Spatio-Temporal Feature Normalization (STFN) on the differences in node features. Clearly, the normalized variance $\hat{\sigma}$ is constrained by the normalization configuration. In cases where $\sigma$ is large ($\sigma > 1$), it holds that $\sigma^{(1-\frac{1}{p})} < \sigma$. Therefore, STFN can reduce the variance of node features. Moreover, the smaller the value of $p$, the more significant the reduction effect, with the strictest constraint observed when $p=1$. In the experimental analysis, setting $p$ to 2 has yielded desirable results.

In addition to adjusting the differences in node feature distributions, STFN can further reduce the neural activity of the network, as shown in Figure \ref{fr}. Regardless of the presence of a residual structure, and whether it is within each layer of a ten-layer model or within models of different depths, STFN consistently lowers the firing rates. This implies that irrelevant neural activity is suppressed, making it more conducive to low-power on-chip inference.

\subsection{Efficiency Evaluation}

\begin{figure*}[h]
\begin{center}
\includegraphics[width=12.0cm]{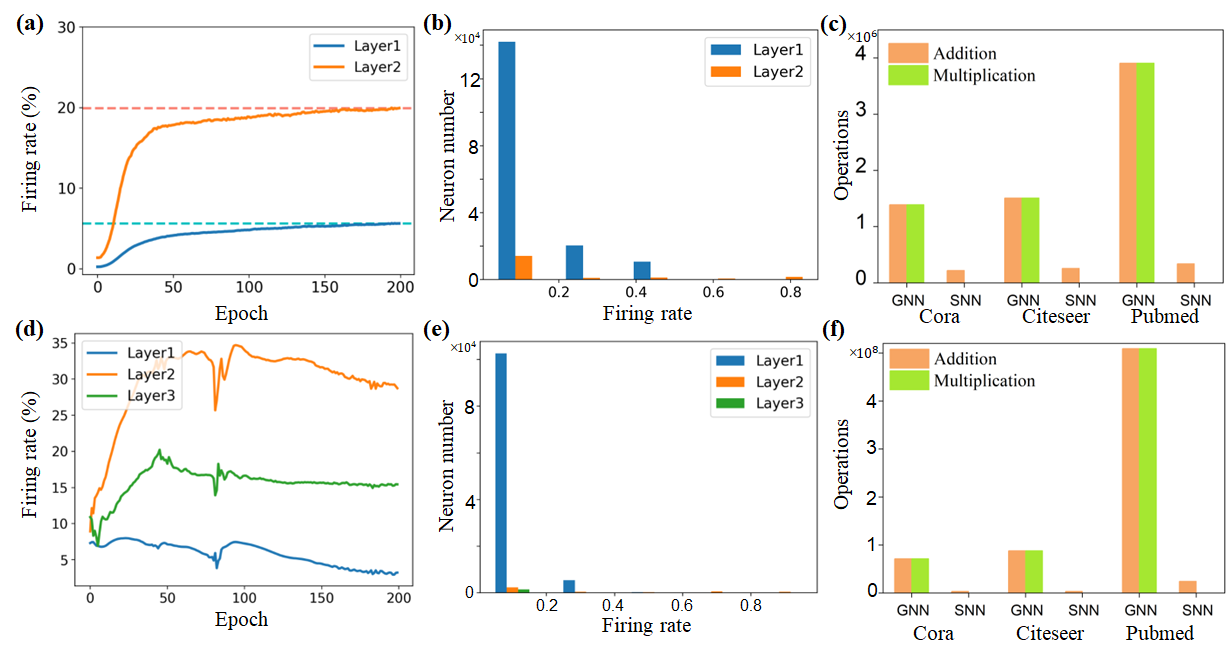}   
\end{center}
\caption{(a)(d) Firing rate records during the training process of two-layer and three-layer SNNs; (b)(e) Statistics of neuron firing rate distribution after training for two-layer and three-layer SNNs; (c)(f) Comparison of addition and multiplication operations between two-layer or three-layer SNNs and GNN.
}
\label{cost}
\end{figure*}

In this section, we further evaluate the computational cost of the proposed spiking neural network for graph learning. As shown in Figure $\ref{cost}$, it can be observed that the neural activity intensity of SNN in graph learning tasks is relatively low during the training process. Since the second layer's features are used for decoding and mapping decisions in this study, the neuron activity density is much higher than that of the first layer. For a two-layer network, the highest average firing rate does not exceed $23\%$ throughout the training process, and for a three-layer network, the highest firing rate does not exceed $27\%$. The majority of neurons have firing rates below $10\%$ (see Figure $\ref{cost}$(b)(e)). This confirms the highly sparse nature of pulse activity within each layer. Overall, even with different network structures, the sparse characteristics of the activity are consistently maintained. We measured the firing rates of trained SNN models with different numbers of layers (SNN-2L, SNN-3L) on Cora, Citeseer, and Pubmed. As shown in Table $\ref{cost-t1}$, all models with different numbers of layers maintain low firing rates and significant sparsity on these datasets.

\begin{table}
\caption{Comparison of the average firing rate for each layer in the network.}
\centering
\begin{tabular}{lccc}
\toprule
Model$\backslash$Dataset  &  Cora  &  Pubmed  & Citeseer \\
\midrule
SNN-2L(Layer 1)   & 0.0398 & 0.0563  &  0.0212  \\
SNN-2L(Layer 2)   & 0.1087 & 0.1995  &  0.0829  \\
\midrule
SNN-3L(Layer 1)   & 0.0342 & 0.0454  &  0.0197  \\
SNN-3L(Layer 2)   & 0.1451 & 0.1740  &  0.0971  \\
SNN-3L(Layer 3)   & 0.1157 & 0.2265  &  0.0922  \\
\bottomrule
\end{tabular}
\label{cost-t1}
\end{table}

This paper compares the computational costs of affine transformations in GNN and SNN networks across three datasets. As shown in Figure \(\ref{cost}(c)(f)\), the SNN model involves fewer operations, with significantly fewer multiplication operations than the GNN model. This results in compression ratios ranging from \(11.62\times\) to \(23.00\times\) (comparing GNN to SNN operations for feature mapping). Table \(\ref{cost-t2}\) indicates that for a three-layer network, the computational compression ratio increases to between \(39.46\times\) and \(53.61\times\). This computational advantage is expected to become more pronounced in deeper and larger network structures.


\begin{table}
\caption{Operation comparison on benchmark datasets. $\ast$ denotes the compression ratio (GNN Opts. / SNN Opts.) in feature transformation process.}
\centering
\begin{tabular}{lccc}
\toprule
Operations($\times10^6$)  &  Cora  &  Pubmed  & Citeseer \\
\midrule
GNN(2 layers)     & 2.78  & 7.82   &  3.02 \\
SNN(2 layers)     & 0.22  & 0.34   &  0.26 \\
GNN(3 layers)     & 143.24   & 1018.66    &  175.84   \\
SNN(3 layers)     & 3.63   & 24.36    &  3.28   \\
\midrule
Compre. ratio(2 layers) $^\ast$  & 12.62$\times$ & 23.00$\times$ & 11.62$\times$  \\ %
Compre. ratio(3 layers) $^\ast$  & 39.46$\times$ & 41.82$\times$  &  53.61$\times$ \\ 
\bottomrule
\end{tabular}
\label{cost-t2}
\end{table}

We compare the affine transformation costs between GNN and SNN networks, both with a 2-layer structure, across three datasets. As depicted in Figure \(\ref{cost}\), our SNN models require no additional operations and significantly fewer multiplication operations than GNN models, achieving a compression ratio between \(11.62\times\) and \(23.00\times\). In Table $\ref{cost-t2}$, this advantage becomes more pronounced with deeper and larger network structures, with the compression ratio increasing to between \(39.46\times\) and \(53.61\times\). 

The efficiency advantage of SNNs is fundamentally due to spike communication and event-driven computation. Binary spiking features allow dense matrix operations to be converted from multiplication to addition, which consumes less energy on neuromorphic hardware. Moreover, the event-driven nature of SNNs enables sparse firing and representation of node features, significantly reducing operational costs. Neuromorphic hardware, which uses distributed memory and self-contained computational units in a decentralized many-core architecture, can fully exploit spike-based communication and local computation. Our model thus provides a template for advancing graph applications in neuromorphic computing.

\section{Conclusion}

In this study, we present a general SNN framework designed for graph-structured data, employing an iterative spiking message passing method. Introducing the Spatio-Temporal Feature Normalization (STFN) method, we merge graph propagation and spiking dynamics into a unified paradigm, reconciling them in a collaborative manner. Our framework is adaptable and applicable to various propagation operations and scenarios, exemplified through instantiation into two specific models: GC-SNN for graph convolution and GA-SNN for graph attention. Experimental results on three benchmark datasets showcase the effectiveness of the models and their robust representation capabilities. Notably, Graph SNNs exhibit high-efficiency advantages, making them well-suited for implementation on neuromorphic hardware and applications in graphic scenarios. Overall, this work contributes novel insights into the intersection of spiking dynamics and graph topology, enhancing our understanding of advanced cognitive intelligence.

\section*{Acknowledgements}

\bibliography{mybibfile}

\end{document}